\documentclass[fleqn,10pt]{wlscirep}
\usepackage[utf8]{inputenc}
\usepackage[T1]{fontenc}

\makeatletter
\let\c@lofdepth\relax
\let\c@lotdepth\relax
\makeatother
\usepackage{xspace}
\usepackage{subfigure}
\usepackage{caption}
\usepackage{booktabs}
\usepackage{silence}
\usepackage{float}
\usepackage{amsmath}
\usepackage[mathlines]{lineno}
\usepackage{tabularx} 
\usepackage{ragged2e}
\usepackage{booktabs} 
\usepackage{colortbl}
\usepackage{multirow}

\title{A Novel Dual-pooling Attention Module for UAV Vehicle Re-identification}

\author[1]{Xiaoyan Guo}
\author[1]{Jie Yang}
\author[1]{Xinyu Jia}
\author[1]{Chuanyan Zang}
\author[1,*]{Yan Xu}
\author[2]{Zhaoyang Chen}

\affil[1]{College of Electronic and Information Engineering, Shandong University of Science \& Technology, Qingdao, 266590, China}
\affil[2]{College of Mathematics and Physics, Beijing University of Chemical Technology, Beijing, 100029, China}
\affil[*]{Email: xuyan@sdust.edu.cn}

\begin{abstract}
Vehicle re-identification (Re-ID) involves identifying the same vehicle captured by other cameras, given a vehicle image. It plays a crucial role in the development of safe cities and smart cities. With the rapid growth and implementation of unmanned aerial vehicles (UAVs) technology, vehicle Re-ID in UAV aerial photography scenes has garnered significant attention from researchers. However, due to the high altitude of UAVs, the shooting angle of vehicle images sometimes approximates vertical, resulting in fewer local features for Re-ID. Therefore, this paper proposes a novel dual-pooling attention (DpA) module, which achieves the extraction and enhancement of locally important information about vehicles from both channel and spatial dimensions by constructing two branches of channel-pooling attention (CpA) and spatial-pooling attention (SpA), and employing multiple pooling operations to enhance the attention to fine-grained information of vehicles. Specifically, the CpA module operates between the channels of the feature map and splices features by combining four pooling operations so that vehicle regions containing discriminative information are given greater attention. The SpA module uses the same pooling operations strategy to identify discriminative representations and merge vehicle features in image regions in a weighted manner. The feature information of both dimensions is finally fused and trained jointly using label smoothing cross-entropy loss and hard mining triplet loss, thus solving the problem of missing detail information due to the high height of UAV shots. The proposed method's effectiveness is demonstrated through extensive experiments on the UAV-based vehicle datasets VeRi-UAV and VRU.
\end{abstract}
\begin{document}

\flushbottom
\maketitle
%
%
\thispagestyle{empty}


\section*{Introduction}

As an important component of intelligent transportation systems, vehicle re-identification (Re-ID) aims to find the same vehicle from the vehicle images taken by different surveillance cameras. The use of vehicle Re-ID algorithm can automatically perform the work of image matching, solving the problem of vehicle identification due to the influence of external conditions, such as artificially blocked license plates, obstacle blocking, blurred images, etc., saving manpower and consuming less time, providing strong technical support for the construction and maintenance of urban security order and guaranteeing public safety. Driven by deep learning technology, more and more researchers have started to shift towards the deep convolutional neural network, which solves the previous problem of insufficient feature extraction expression using traditional methods. 

\begin{figure*}[t]
\centering 
\subfigure[Road surveillance-based image display]{
\label{Fig.sub.1}
\includegraphics[width=8.6cm,height = 2.8cm]{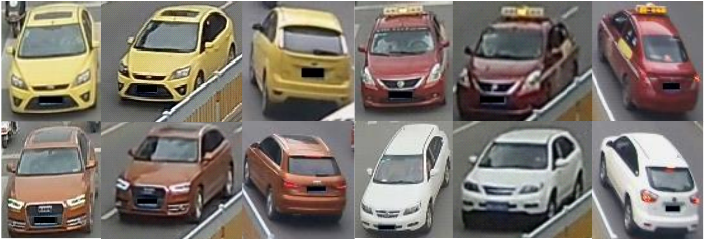}}
\subfigure[UAV-based image display]{
\label{Fig.sub.2}
\includegraphics[width=8.6cm,height = 2.8cm]{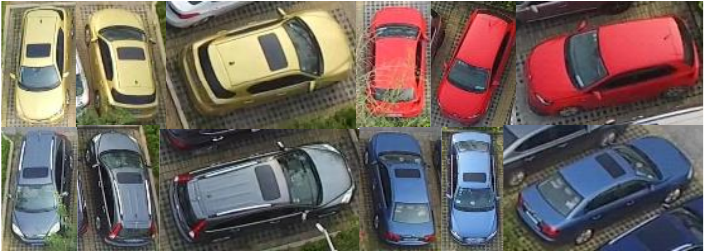}}
\caption{Comparison of two types of vehicle images.}
\label{fig1}
\end{figure*}

Existing vehicle Re-ID work~\cite{zhu2019vehicle,shen2021exploring,he2020multi,zheng2020vehiclenet,rong2021vehicle,shen2023triplet} is mainly through road surveillance video to obtain vehicle data. A large number of surveillance cameras deployed in highways, intersections and other areas can only provide a specific angle and a small range of vehicle images. When encountering certain special circumstances, such as camera failure or events that the target vehicle is not in the monitoring coverage,  it is impossible to identify and re-identify the target vehicle. In recent years, unmanned aerial vehicles (UAVs) technology~\cite{outay2020applications} has made significant developments in terms of flight time, wireless image transmission, automatic control, etc. Mobile cameras on UAVs have a wider range of viewpoints as well as better maneuverability, mobility, and flexibility, and UAVs can track and record specific vehicles in urban areas and highways~\cite{wang2019development}. Therefore, the vehicle Re-ID task in the UAV scenario has received increasingly wide attention from researchers as a complementary development to the traditional road surveillance scenario and has greater application value in practical public safety management, traffic monitoring, and vehicle statistics. Figure \ref{fig1} compares the two types of vehicle images based on road surveillance and aerial photography based on UAVs. The similarity between the two is that the captured vehicle image is a single complete vehicle. The difference is that the height of the UAV is usually higher than the height of the fixed surveillance camera, which results in the angle of the vehicle image sometimes being approximately vertical. Also, the height of the UAV is uncertain, resulting in scale variation in the captured vehicle images.

Since the height of the UAV is usually higher than the height of the fixed surveillance camera, the obtained vehicle images are taken at an almost near vertical angle, and therefore fewer local features of the vehicle are used for Re-ID. On the one hand, the idea of the attention mechanism has been proven to be effective. It is important to build an attention module to focus on channel information and important regions. On the other hand, average pooling~\cite{lin2013network} takes the average value in each rectangular region, which preserves the background information in the image and allows input of the information of extracting all features in the feature map to the next layer. Generalized mean pooling operation~\cite{gu2018attention} allows focusing on regions with different fineness by adjusting the parameters. The minimum pooling operation~\cite{kauffmann2020towards} will focus on the smallest pixel points in the feature map. Soft pooling~\cite{stergiou2021refining} is based on softmax weighting to retain the basic attributes of the input while amplifying the feature activation with greater intensity, i.e., to minimize the information loss brought about by the pooling process and to better retain the information features. Unlike maximum pooling, soft pooling is differentiable, so the network acquires a gradient for each input during backpropagation, which facilitates better training. A series of pooling methods have been successively proposed by researchers~\cite{zhai2017s3pool,gulcehre2014learned,zhao2021liftpool}, each of which has shown different advantages and disadvantages. Previous studies usually combine only average pooling and maximum pooling to capture key features of images, while ignoring the use of multiple pooling methods in combination. In addition, the pooling layer is an important component in convolutional neural networks and has a significant role in reducing the number of network training parameters, decreasing the difficulty of network optimization, and preventing overfitting~\cite{sun2017learning}. 

\subsection*{Vehicle Re-ID methods}
In recent years, most vehicle Re-ID methods are based on traditional road surveillance images, and their methodological ideas broadly include using vehicle local features to achieve the extraction of detailed feature information~\cite{liu2018ram,chen2019partition,wang2017orientation}, using attention mechanisms to improve the model's ability to focus on important regions~\cite{zhang2021sha,shen2022hsgm,pan2022vehicle}, optimizing network training to improve recognition rates by designing appropriate loss functions~\cite{bai2018group,li2020vehicle}, and using unsupervised learning without manual labeling to improve the generalization ability of the model in complex realistic scenes~\cite{peng2020cross,song2020unsupervised,bashir2019vr}.
For example, Jiang et al.~\cite{jiang2022global} designed a global reference attention network (GRA-Net) with three branches to mine a large number of useful discriminative features to reduce the difficulty of distinguishing similar-looking but different vehicles. Viewpoint-Aware Network (VANet)~\cite{wang2021viewpoint} is used to learn feature metrics for the same and different viewpoints. Generative adversarial networks (GAN) are used to solve the labeling difficulty in the Re-ID dataset~\cite{wu2019vehicle,wang2021inter}.  EMRN~\cite{shen2021efficient} proposes a multi-resolution features dimension uniform module to fix dimensional features from images of varying resolutions, thus solving the multi-scale problem. Besides, GiT \cite{shen2023git} uses a graph network approach to propose a structure where graphs and transformers interact constantly, enabling close collaboration between global and local features for vehicle Re-ID.

However, the current vehicle datasets VeRi-776, VehicleID, etc. are captured by fixed surveillance cameras, and the perspective and diversity of vehicles are insufficient, so the above-mentioned feature extraction methods are only for vehicle images captured by traditional road surveillance. Since the birth of the first vehicle Re-ID dataset VARI~\cite{wang2019vehicle} based on aerial images in 2019, vehicle Re-ID using images captured by UAVs has started to attract the attention of researchers~\cite{shen2022adaptively, yao2021attention, qiao2022vehicle}. For example, in the aerial image scenario, the normalized softmax loss~\cite{qiao2022vehicle} is proposed to increase the inter-class distance and decrease the intra-class distance and combine with the triplet loss to train the model.

\subsection*{Attention mechanism}
The attention mechanism~\cite{li2022enhancing,zheng2023dual} is widely used in various fields of deep convolutional neural networks, and its core idea is to select the most important information for the current target task from a large amount of information. Channel attention squeeze-and-excitation (SE) network~\cite{hu2018squeeze} plays the role of emphasizing important channels while suppressing noise. Spatial attention non-local~\cite{wang2018non} considers the weighting of all location features to obtain more comprehensive semantic information. Triplet attention~\cite{misra2021rotate}, which predicts channel and spatial attention separately, considers the relationship between two neighborhoods through three branches to achieve cross-domain interaction.

Recently, many researchers have combined the attention mechanism with vehicle Re-ID models to substantially improve the feature representation capability of the models. For example, the dual-relational attention module (DRAM)~\cite{zheng2023dual} models the importance of feature points in the spatial dimension and the channel dimension to form a three-dimensional attention module to improve the performance of the attention mechanism and mine more detailed semantic information. In addition, Zhang et al.~\cite{zhang2022dual} proposed a dual-attention granularity network approach for vehicle reconstruction, which used an embedded self-attention model to obtain an attentional heat map and then obtained accurate local regions by partial localization on the attentional heat map. However, unlike traditional road surveillance cameras, the height of UAV aerial photography is more flexible and usually higher than that of fixed surveillance cameras, resulting in captured vehicle images that are mostly non-complete vehicles captured in a top-down view. Consequently, some generic attention modules cannot capture local detail regions well, resulting in poor performance of vehicle Re-ID models based on UAV aerial images.

Based on the above analysis, this paper presents a novel dual-pooling attention (DpA) module for UAV vehicle Re-ID. Specifically, we first design the channel-pooling attention (CpA) module and the spatial-pooling attention (SpA) module respectively, by combining multiple pooling operations, so that the network can better focus on detailed information while avoiding the intervention of more redundant information, and the pooling operations are also useful in preventing overfitting. Second, the CpA module is designed to focus on the important features of the vehicle while ignoring unimportant information. The SpA module is designed to capture the local range dependence of spatial regions, and then the two are concatenated to obtain the dual-pooling attention module, which is embedded into the conventional ResNet50 backbone network to improve the model's perception of channels and spaces. Among them, omni-dimensional dynamic (OD) convolution is also introduced in the CpA and SpA modules to further extract rich contextual information dynamically. In addition, the paper introduces hard mining triplet loss combined with cross-entropy loss with label smoothing for training, thus improving the ability of triplet loss to perform strong discrimination even in the face of difficult vehicle samples. Finally, a large number of experiments are conducted to verify the effectiveness of our model, and the results show that the proposed method can achieve an mAP of 81.74\% on the VeRi-UAV dataset. On the three test subsets of VRU, the accuracy of mAP reaches 98.83\%, 97.90\%, and 95.29\%, respectively. This indicates that the DpA module can solve the problem of insufficient fine-grained information based on vehicle Re-ID images taken by UAVs.

\section*{Methods}
\subsection*{Overall network architecture}
\begin{figure}[t]
\centering
\includegraphics[width=17.5cm,height=6cm]{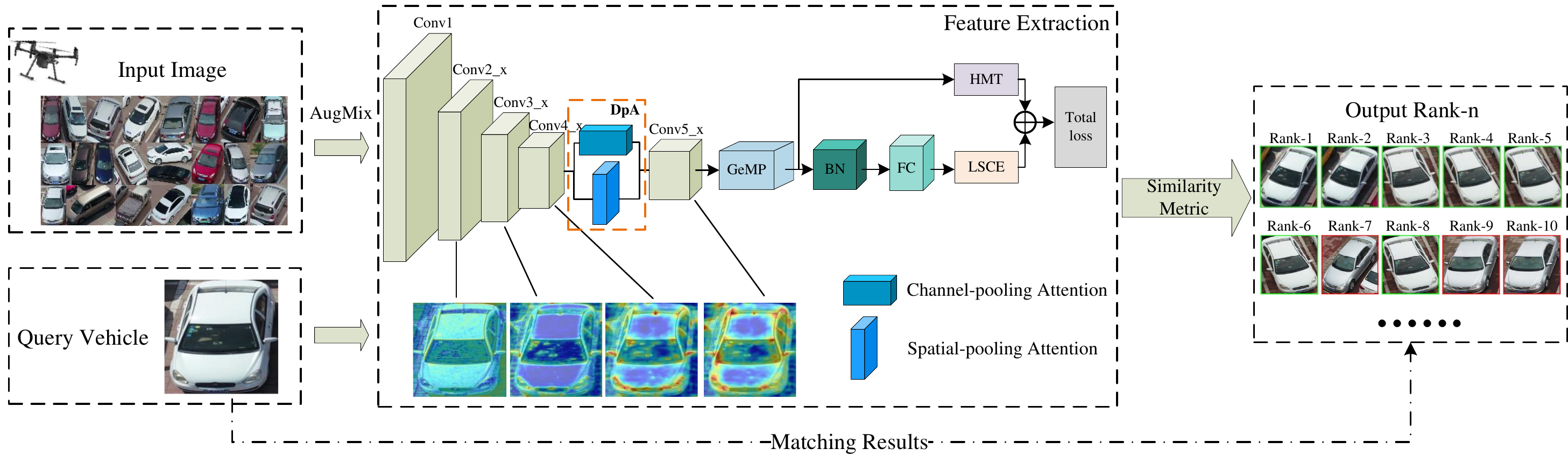}
\caption{The overall framework of the network for vehicle Re-ID.\label{fig2}}
\end{figure}

The overall network architecture of this paper is shown in Figure \ref{fig2}. It consists of three parts: input images, feature extraction, and output results. First, the input image is enhanced with data by AugMix~\cite{hendrycks2019augmix} method, where AugMix overcomes the image distortion problem caused by previous MixUp data enhancement by applying different data enhancements randomly to the same image. Then, the backbone network ResNet50 and a dual-pooling attention (DpA) module are used as the feature extraction part of the network. After the gallery set to be queried and the target query vehicle are input to the network model for feature extraction, the similarity between the features of the target query vehicle image and the vehicle image features in the gallery set is calculated by a metric method. Finally, the similarity is ranked and the vehicle retrieval results are obtained.

\subsection*{Channel-pooling attention}
\begin{figure*}[t]
    \centering
    \includegraphics[width=1\textwidth]{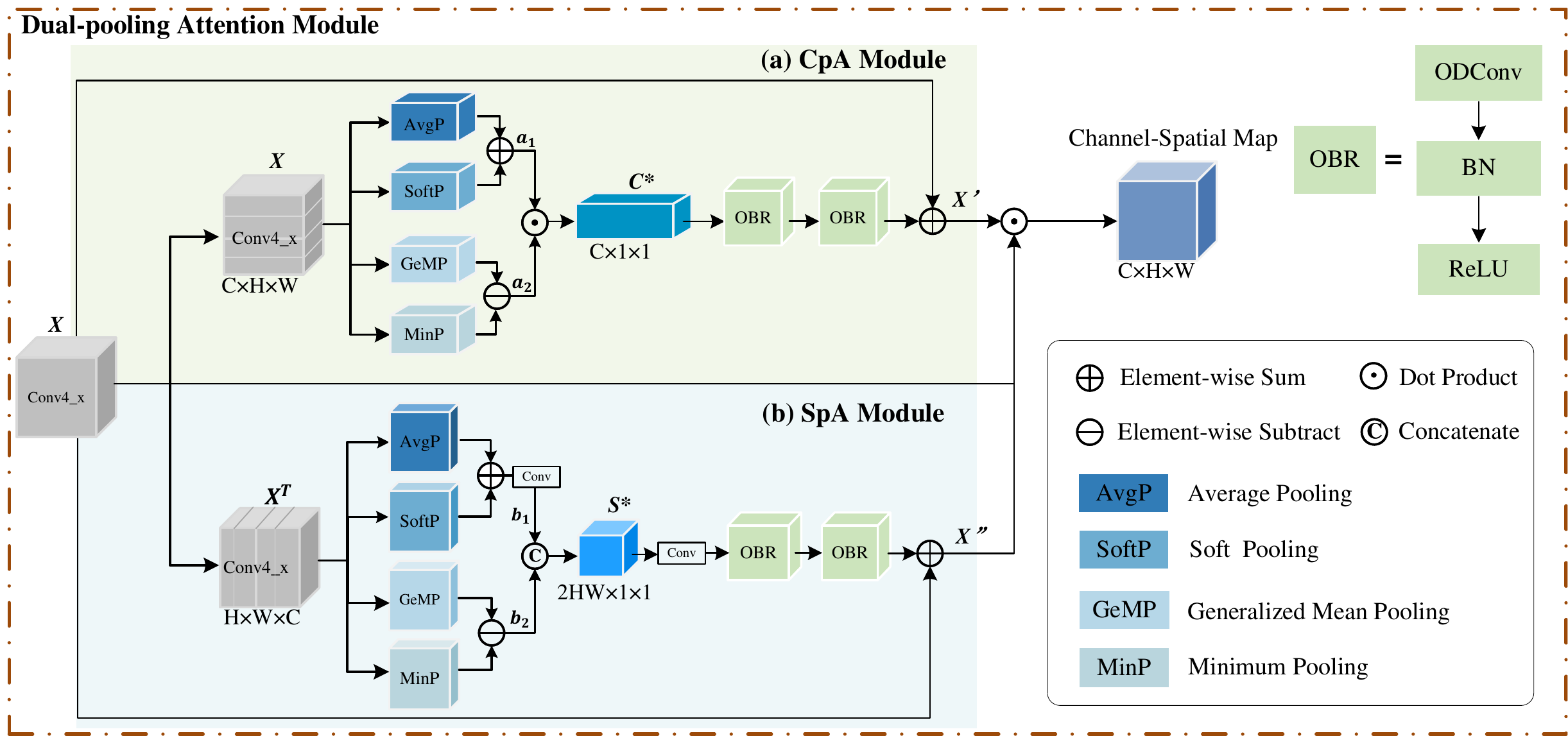}
    \caption{{Dual-pooling attention module. (a) CpA represents channel-pooling attention, (b) SpA represents spatial-pooling attention.}}
    \label{fig3}
\end{figure*}

To focus more on the features with the discriminative nature of vehicle images and avoid the interference of background clutter information, four pooling methods are introduced to process the channel features. The specific module diagram is shown in Figure \ref{fig3} (a).
First, let the output features of the third residual block (Conv4\_x) of Resnet50 be the input matrix $X$. Suppose the input matrix $\emph{X}\in\mathbb{R}^{C\times H\times W}$, where $C$, $H$, and $W$ represent the channel number, height, and width of the feature map respectively. Four copies of $X$ are made, and the average pooling (AvgP)~\cite{lin2013network}, generalized mean pooling (GeMP)~\cite{gu2018attention}, minimum pooling (MinP)~\cite{kauffmann2020towards}, and soft pooling (SoftP)~\cite{stergiou2021refining} operations are performed on them. The first three poolings make the dimension change from $ C\times H\times W$ to $C\times 1\times 1$ channel descriptors. The feature map $\emph{X}\in \mathbb{R}^{C\times H\times W}$ is taken as input and a vector $\emph{f}\in \mathbb{R}^{C\times 1\times 1}$ is generated as the output of the pooling operation. The vector $f=[f^1,...f^k,...f^C]$ in the case of the AvgP, MinP and GeMP of are respectively given by:

\begin{equation}\label{eqs1}
f_{AvgP}^{k}=\frac{1}{|R_{ij}|}\sum_{(p,q)\in R_{ij}}x_{pq}^{k}
\end{equation}

\begin{equation}\label{eqs3}
f_{MinP}^{k}=-\max_{(p,q)\in R_{ij}}(-x_{pq}^{k})
\end{equation}
where $x_{pq}^{k}$ denotes the element located at $(p,q)$ in the rectangular region $R_{ij}$, $|R_{ij}|$ indicates the number of elements in the rectangular area $R_{ij}$.

\begin{equation}\label{eqs2}
f_{GeMP}^{k}=\left(\frac{1}{|R|}\sum_{(p,q)\in R}{(x_{pq}^{k})^\alpha}\right)^\frac{1}{\alpha}
\end{equation}
where $x_{pq}^{k}$ denotes the element located at $(p,q)$ in the rectangular region $R$, $|R|$ denotes the number of all elements of the $k_{th}$ feature map, and $\alpha$ is the control coefficient.

And the feature map generated by SoftP is still $C\times H\times W$. Its formulas for SoftP are shown as follows:
\begin{equation}\label{eqs4}
f_{SoftP}^{k}=\frac{1}{\sum_{(m,n)\in R}e^{x_{mn}^{k}}}\sum_{(p,q)\in R}e^{x_{pq}^{k}}\times x_{pq}^{k}
\end{equation}
where $x_{mn}^{k}$ is similar to $x_{pq}^{k}$ above and denotes the element located at $(m,n)$ in the rectangular region $R$.

From one perspective, since AvgP focuses on each pixel of the feature map equally and SofP captures important regions better than maximum pooling, the outputs of both are summed to obtain $\emph{$a_1$}\in \mathbb{R}^{C\times H\times W}$ to give more attention to important vehicle features. From another perspective, GeMP can focus on different fine-grained regions adaptively by adjusting the parameters, while minimum pooling focuses on small pixels in the feature map, i.e., the background regions, so GeMP and MinP are subtracted to obtain $\emph{$a_2$}\in \mathbb{R}^{C\times 1\times 1}$ to give more attention to vehicle fine-grained features and ignore the background regions as much as possible. The output of both of them is dotted and multiplied to obtain the channel attention map $\emph{$C^*$}\in \mathbb{R}^{C\times H\times W}$. The channel pooling matrix $\emph{$C^*$}$ can be formulated as:
\begin{equation}\label{eqs5}
a_1=Conv(AvgP(X)+SoftP(X))
\end{equation}
\begin{equation}\label{eqs6}
a_2=GeMP(X)-MinP(X)
\end{equation}
\begin{equation}\label{eqs7}
C^*=a_1*a_2
\end{equation}
where Conv stands for convolution operation and * represents the dot product operation.

The OBR module is composed of OD convolution, batch normalization (BN) and rectified linear unit (ReLU) activation function, which is sequentially used twice in a row for the channel attention map $C^*$. Compared with normal convolution, dynamic convolution is used here, which is linearly weighted by multiple convolution kernels and establishes certain dependencies with the input data to better learn flexible attention and enhance the extraction of feature information. Finally, the original input matrix $X$ is summed with the output of the OBR module and normalized by the sigmoid function to obtain the final channel-pooling attention output matrix $ X^{'}\in \mathbb{R}^{C\times H\times W}$. These operations can be defined as:

\begin{equation}\label{eqs8}
X^{'}=\sigma(OBR(OBR(C^*))+X)
\end{equation}
where $\sigma$(·) is the sigmoid activation function and the OBR module represents the OD convolution of 3$\times$3, BN, and ReLU activation function.

\subsection*{Spatial-pooling attention}
Feature relations are used to compute spatial attention, similar to the above channel-pooling attention module. As shown in Figure \ref{fig3} (b), first, the output feature $X$ of the original feature, the third residual block of Resnet50 (Conv4\_x), is transposed to obtain $X^{T}\in \mathbb{R}^{H\times W\times C}$. Then the operation of multiplying $H$ and $W$ is performed to aggregate and extend the dimensions to become a matrix of $HW\times C\times 1$. This matrix is then copied in four copies and AvgP, SoftP, GeMP, and MinP are applied along the channel axis, which finally makes the dimension change from $H\times W \times C$ to $HW\times 1\times 1$ spatial descriptors. Similarly, the outputs of AvgP and SoftP are added and the convolution layer is applied to obtain $b_1\in \mathbb{R}^{HW\times 1\times 1}$. The outputs of GeMP and MinP are subtracted to obtain $b_2\in \mathbb{R}^{HW\times 1\times 1}$. Finally, the two are concatenated to get the output $S^*\in \mathbb{R}^{2HW\times 1\times 1}$. The spatial pooling matrix \emph{$S^*$} can be formulated as:
\begin{equation}\label{eqs9}
b_1=Conv(AvgP(X^{T})+SoftP(X^{T}))
\end{equation}
\begin{equation}\label{eqs10}
b_2=GeMP(X^{T})-MinP(X^{T})
\end{equation}
\begin{equation}\label{eqs11}
S^*=[b_1,b_2]
\end{equation}
where [· , ·] is the concatenation operation.

Then convolution is applied to $S^*$ to expand it to $C\times 1\times 1$. Similarly, the OBR module uses twice for the output attention map $S^*$ to dynamically enhance the acquisition of spatial domain information features. Finally, the original input $X$ is added up to get the output matrix $X^{''}\in \mathbb{R}^{C \times H \times W}$ of the spatial set attention module. These operations can be defined as:
\begin{equation}\label{eqs12}
X^{''}=\sigma(OBR(OBR(Conv(S^*)))+X)
\end{equation}

\subsection*{Loss functions}
In vehicle Re-ID, a combination of identity loss and metric loss is often used. Therefore, in the training phase, we use cross-entropy (CE) loss for classification and triplet loss for metric learning. The CE loss is often used in classification tasks to represent the difference between the true and predicted values. The smaller the value, the better the prediction of the model. The label smoothing (LS) strategy~\cite{szegedy2016rethinking} is introduced to solve the overfitting problem. Therefore, the formula for the label smoothing cross-entropy (LSCE) loss is as follows:
\begin{equation}\label{eqs13}
\mathcal{L}_{LSCE}=\begin{cases}
-\sum\limits_{i=1}^{N}({1-\frac{N-1}{N}{\varepsilon}})\log(p_{i}),\quad &n=i \\
-\sum\limits_{i=1}^{N}\frac{N}{\varepsilon}\log(p_{i}),\quad &n\neq i
\end{cases} 
\end{equation}
where parameter $\varepsilon$ is the smoothing factor, which was set to 0.1 in the experiment.

The core idea of triplet loss is to first build a triplet consisting of anchor samples, positive samples, and negative samples. Then after continuous learning, the distance between positive samples and anchor samples under the same category in the feature space is made closer, and the distance between negative samples and anchor samples under different categories are made farther. In this paper, we use hard mining triplet (HMT) loss to further improve the mining ability in the face of difficult vehicle samples by selecting the more difficult to distinguish positive and negative sample pairs in a batch for training. The loss function for the hard mining triplet is calculated as follows:

\begin{linenomath}
\begin{equation}\label{eqs14}
\mathcal{L}_{HMT}=\sum\limits_{i=1}^{T}\sum\limits_{A=1}^{S}[\max\limits_{P=1,...,T}(d(A_i,P_i))-\min_{\mbox{\tiny$\begin{array}{c}
     N=1,...,S\\
      j=1,...,T\\
      i\neq j\end{array}$}}(d(A_i,N_j))+m]_+ 
\end{equation}
\end{linenomath}
where $T$ denotes the number of vehicle identities in each training batch, $S$ denotes the number of images included in each vehicle identity. $A_i$, $P_i$, and $N_j$ denote the anchor sample, the vehicles belonging to the same category as the anchor sample but least similar to it, and the vehicles belonging to a different category than the anchor sample but most similar to it, respectively. $m$ represents the minimum boundary value of this loss, and [·]$_+$ is the max(·, 0) function.

In summary, this work combines LSCE loss and HMT loss. The final loss is:
\begin{equation}\label{eqs15}
\mathcal{L}_{total}=\lambda_1\mathcal{L}_{LSCE}+\lambda_2\mathcal{L}_{HMT}
\end{equation}
where $\lambda_1$ and $\lambda_2$ are two weights for different losses, and $\lambda_1$ = $\lambda_2$ = 1.

\section*{Experiment results and discussion}
\subsection*{Datasets}
Song et al.~\cite{liu2022posture} constructed \textbf{VeRi-UAV}, a dataset based on the Re-ID of UAV vehicles, to capture vehicles from multiple angles in different areas, including parking lots and highways. VeRi-UAV includes 2,157 images of 17,516 complete vehicles with 453 IDs. To test the Re-ID method, the authors segmented another 17,516 vehicle images using a vehicle segmentation model. After some minor manual adjustments, the dataset has a total of 9,792 training images, 6,489 test images, and 1,235 query images. Lu et al.~\cite{lu2022vehicle} constructed \textbf{VRU}, the largest current vehicle Re-ID dataset based on aerial drone photography. The dataset was divided into a training set and three test sets: small, medium, and large. The training set includes 80,532 images of 7,085 vehicles. The small, medium, and large test sets contain 13,920 images of 1,200 vehicles, 27,345 images of 2,400 vehicles, and 91,595 images of 8,000 vehicles, respectively.

\subsection*{Implementation details}
In this paper, we use the weight parameters of ResNet50 pre-trained on ImageNet as the initial weights of the network model. All experiments were performed on PyTorch. For each training image, balanced identity sampling is taken and it is resized to $256\times 256$ and pre-processing is also performed using the AugMix data augmentation method. In the training phase, the model was trained for a total of 60 epochs, and a warm-up strategy using a linear learning rate was employed. 

For the VeRi-UAV dataset, the training batch size is 32 and an SGD optimizer was used with an initial learning rate of 0.35e-4. The learning rate tuning strategy of CosineAnnealingLR is also used. For the VRU dataset, the training batch size is 64 and the initial learning rate is 1e-4 using the Adam optimizer. The learning rate tuning strategy of MultiStepLR is used, which decays to 1e-5 and 1e-6 in the 30th and 50th epochs. In addition, the batch sizes for testing are all 128. In the model testing phase, we use Rank-n and mean average precision (mAP) as the main evaluation metrics. The mINP is also introduced in the ablation experiments to further demonstrate the experimental effects.

\subsection*{Comparison with state-of-the-art methods}
\emph{Comparisons on VeRi-UAV.}
The methods compared on the VeRi-UAV dataset include the handcrafted feature-based methods BOW-SIFT~\cite{liu2016large} and LOMO~\cite{liao2015person}, and the deep learning-based methods Cross-entropy Loss~\cite{szegedy2016rethinking}, Hard Triplet Loss~\cite{hermans2017defense}, VANet~\cite{chu2019vehicle}, Triplet+ID Loss~\cite{wang2019vehicle}, RANet~\cite{yang2020resolution}, ResNeSt~\cite{zhang2022resnest} and PC-CHSML~\cite{liu2022posture}. Among them, LOMO~\cite{liao2015person} improves vehicle viewpoint and lighting changes through handcrafted local features. BOW-SIFT~\cite{liu2016large} performs feature extraction by employing content-based image retrieval and SIFT. VANet~\cite{chu2019vehicle} learns visually perceptive depth metrics and can retrieve images with different viewpoints under similar viewpoint image interference. RANet~\cite{yang2020resolution} implements a deep CNN to perform resolution adaptive. PC-CHSML~\cite{liu2022posture} are approaches for UAV aerial photography scenarios, which improves the recognition retrieval of UAV aerial images by combining pose-calibrated cross-view and difficult sample-aware metric learning. Table \ref{tab1} shows the comparison results with the above-mentioned methods in detail. First of all, the results show that the deep learning-based approach achieves superior improvement over the manual feature-based approach. Secondly, compared with methods for fixed surveillance shooting scenarios such as VANet~\cite{chu2019vehicle}, DpA shows some improvement in shooting highly flexible situations. Additionally, compared with the method PC-CHSML~\cite{liu2022posture} for the UAV aerial photography scenario, DpA shows an improvement of 4.2\%, 10.0\%, 9.5\%, and 9.6\% for different metrics of mAP, Rank-1, Rank-5, and Rank-10. Consequently, the effectiveness of the module is further verified.

\begin{table}[t]
\centering
\begin{tabular}{|l|c|c|c|c|}
\hline
\rowcolor{gray}
\textbf{Method}	& \textbf{mAP}	& \textbf{Rank-1} & \textbf{Rank-5}& \textbf{Rank-10}\\
\hline
BOW-SIFT~\cite{liu2016large} & 6.7   & 18.9  & 34.4 & 43.4   \\
\hline
LOMO~\cite{liao2015person} & 25.5  & 51.9   & 70.1 & 77.0    \\
\hline
RANet~\cite{yang2020resolution} & 44.3  & 71.6    & 82.4& 85.3   \\
\hline
ResNeSt~\cite{zhang2022resnest} & 64.4  & 80.1   & 85.5 & 86.9    \\
\hline
Triplet+ID Loss~\cite{wang2019vehicle} & 66.1  & 80.9   & 86.9 & 88.4 \\
\hline
VANet~\cite{chu2019vehicle} & 66.5  & 81.6  & 87.0 & 88.0   \\
\hline
Cross-entropy Loss~\cite{szegedy2016rethinking} & 67.6     &  94.8   & 96.3   &97.4\\
\hline
Hard Triplet Loss~\cite{hermans2017defense} & 73.2  & 84.8 & 88.6 & 89.2 \\
\hline    
PC-CHSML~\cite{liu2022posture} & {77.5}  & 86.6 & 89.0 & 89.8    \\
\hline
\textbf{DpA (Ours)} & \textbf{81.7} & \textbf{96.6} & \textbf{98.5}  & \textbf{99.4} \\
\hline
\end{tabular}
\caption{\label{tab1}Comparison of various proposed methods on VeRi-UAV dataset (in \%). Bold numbers indicate the best ranked results.}
\end{table}

\begin{table*}[t]
  \centering
  \begin{tabular}{|l|c|c|c|c|c|c|c|c|c|}
    \hline
    \rowcolor{gray}
    \multicolumn{1}{|c|}{}       & \multicolumn{3}{c|}{\textbf{Small}} & \multicolumn{3}{c|}{\textbf{Medium}} & \multicolumn{3}{c|}{\textbf{Big}} \\ \cline{2-10} 
    \rowcolor{gray}
    \multicolumn{1}{|l|}{\multirow{-2}{*}{\textbf{Method}}} & \textbf{mAP}    & \textbf{Rank-1}  & \textbf{Rank-5} & \textbf{mAP}    & \textbf{Rank-1}  & \textbf{Rank-5}  & \textbf{mAP}   & \textbf{Rank-1} & \textbf{Rank-5} \\
    \hline
    MGN~\cite{wang2018learning}& 82.48  & 81.72 & 95.08 & 80.06   & 78.75 & 93.75 & 71.53  & 66.25 & 87.15 \\
     \hline
    SCAN~\cite{teng2018scan} & 83.95    & 75.22 & 95.03 & 77.34  & 67.27 & 90.51 & 64.51  & 52.44 & 79.63 \\
     \hline
    Triplet+CE loss~\cite{he2020fastreid} & 97.40 & 95.81 & 99.29 & 95.82  & 93.33 & 98.83 & 92.04   & 87.83 & 97.28 \\
     \hline
    GASNet~\cite{lu2022vehicle} & 98.51  & 97.45 & 99.66 & 97.31 & 95.59 & 99.33 & 93.93  & 90.29 & 98.40 \\
    \hline
    \textbf{DpA(Ours)} & \textbf{98.83}      & \textbf{98.07}      & \textbf{99.70}    & \textbf{97.90}    & \textbf{96.51}      & \textbf{99.44}        & \textbf{95.29}      & \textbf{92.30}       & \textbf{98.96}  \\
    \hline
\end{tabular}
\caption{\label{tab2}Comparison of various proposed methods on VRU dataset (in \%). Bold numbers indicate the best ranked results.}
\end{table*}%

\emph{Comparisons on VRU.} It is a relatively newly released UAV-based vehicle dataset, hence, few results have been reported about it. Table \ref{tab2} compares DpA with other methods~\cite{wang2018learning,teng2018scan,he2020fastreid,lu2022vehicle} on VRU dataset. 
Among them, MGN~\cite{wang2018learning} integrates information with different granularity by designing one global branch and two local branches to improve the robustness of the network model. SCAN~\cite{teng2018scan} uses channel and spatial attention branches to adjust the weights at different locations and in different channels to make the model more focused on regions with discriminative information. Triplet+CE loss~\cite{he2020fastreid} then uses ordinary triplet loss and cross-entropy loss for model training. The GASNet model~\cite{lu2022vehicle} captures effective vehicle information by extracting viewpoint-invariant features and scale-invariant features. The results show that, in comparison, DpA contributes 0.32\%, 0.59\%, and 1.36\% of the mAP improvement to the three subsets of VRU. Taken together, this indicates that the DpA module enhances the ability of the model to extract discriminative features, which can well solve the problem of local features being ignored in UAV scenes.

\subsection*{Ablation experiments}
In this section, we designed some ablation experiments on the VeRi-UAV dataset to evaluate the effectiveness of the proposed methodological framework. The detailed results of the ablation studies are listed in Tables \ref{tab3}, \ref{tab4}, \ref{tab5} and \ref{tab6}. It is worth noting that a new evaluation index mINP was introduced in the experiment. The mINP is a recently proposed metric for the evaluation of Re-ID models i.e., the percentage of correct samples among those that have been checked out as of the last correct result.

\subsubsection*{Evaluation of DpA module}
To verify the validity of the DpA module, we directly used the baseline network composed of ResNet50 as the backbone network, combined with generalized mean pooling, batch normalization layer, fully connected layer, LSCE loss, and HMT loss. The detailed results of the ablation study on the VeRi-UAV dataset are shown in Table \ref{tab3}. Firstly, the results showed that the addition of CpA to the baseline resulted in a 1.62\% and 0.36\% improvement in the assessment over the baseline on mAP and Rank-1, respectively. This indicates that CpA enhances the channel information to be able to extract discriminative local vehicle features. Then after adding SpA to the baseline alone, it improved by 0.67\% and 0.54\% over the baseline on mAP and Rank-1 respectively, showing a greater focus on important regions in the spatial dimension. Finally, after combining CpA and SpA on top of the baseline, we can find another 2.49\%, 0.63\%, 0.36\%, and 2.27\% improvement on mAP, Rank-1, Rank-5, and mINP, respectively. 
We can draw two conclusions: firstly, feature extraction from two dimensions, channel and spatial, respectively, can effectively extract more and discriminative fine-grained vehicle features. Secondly, the accuracy of Re-ID is improved by connecting two attention modules in parallel.

\begin{table}[t]
\centering
\begin{tabular}{|l|c|c|c|c|}
\hline
\rowcolor{gray}
\textbf{Method} &\textbf{mAP}	&\textbf{Rank-1} &\textbf{Rank-5} &\textbf{mINP} \\
\hline
 Resnet50+LSCE+HMT (Baseline)   & 79.25  & 95.96   & 98.12  & 49.29  \\
 \hline
 Baseline+CpA &  \underline{80.87}   &  96.32    & \underline{98.21} &  \underline{49.88} \\
\hline
 Baseline+SpA &  79.92      &  \underline{96.50}   & \underline{98.21} &  48.97\\
\hline
 \textbf{Baseline+DpA} &\textbf{81.74}     & \textbf{96.59}      &  \textbf{98.48}    & \textbf{51.56 }\\
\hline
\end{tabular}
\caption{\label{tab3}Ablation experiments of DpA module on VeRi-UAV (in \%). Bold and underlined numbers indicate the best and second best ranked results, respectively.}
\end{table}

\subsubsection*{Comparison of different attention modules}
This subsection compares the performance with the already proposed attention modules SE~\cite{hu2018squeeze}, Non-local~\cite{wang2018non}, CABM~\cite{woo2018cbam}, and CA~\cite{hou2021coordinate}. Correspondingly, SE~\cite{hu2018squeeze} gives different weights to different positions of the image from the perspective of the channel domain through a weight matrix to obtain more important feature information. Non-local~\cite{wang2018non} achieves long-distance dependence between pixel locations, thus enhancing the attention to non-local features. CBAM~\cite{woo2018cbam} module sequentially infers the attention map along two independent channel and spatial dimensions and then multiplies the attention map with the input feature map to perform adaptive feature optimization. CA~\cite{hou2021coordinate} decomposes channel attention into two one-dimensional feature encoding processes that aggregate features along both vertical and horizontal directions to efficiently integrate spatial coordinate information into the generated attention maps.

Table \ref{tab4} shows the experimental comparison results for different attentional mechanisms. Firstly, adding the SE and CA attention modules can slightly improve the accuracy of the model to some extent, while adding the Non-local, CBAM attention module does not produce the corresponding effect. Second, compared with the newer attention module CA, the proposed DpA module can achieve 2.16\% mAP, 0.36\% Rank-1, 0.27\% Rank-5, and 4.55\% mINP gains on VeRi-UAV. Therefore, this demonstrates the proposed DpA module is more robust in UAV aerial photography scenarios with near-vertical shooting angles and long shooting distances.

\begin{table}[t]
\centering
\begin{tabular}{|l|c|c|c|c|}
\hline
\rowcolor{gray}
\textbf{Method} &\textbf{mAP}	&\textbf{Rank-1} &\textbf{Rank-5} &\textbf{mINP} \\
\hline
  Resnet50+LSCE+HMT (Baseline)   & 79.25       & 95.96   & 98.12  &49.29 \\
\hline
  Baseline+SE~\cite{hu2018squeeze} &  \underline{79.89}      & 96.14      & \textbf{98.48}   &\underline{49.71}    \\
\hline
  Baseline+Non-local~\cite{wang2018non} & 78.85   &  \underline{96.50}   &  \underline{98.21}        &47.42 \\
\hline
  Baseline+CBAM~\cite{woo2018cbam} & 78.69       &  {96.23}      &  97.85    & 48.17  \\
\hline 
  Baseline+CA~\cite{hou2021coordinate} & 79.58     & 96.23       & \underline{98.21}     &47.01\\
\hline
 \textbf{Baseline+DpA} & \textbf{81.74}    & \textbf{96.59}      &  \textbf{98.48}     & \textbf{51.56}\\
\hline
\end{tabular}
\caption{\label{tab4}Ablation experiments of different attention modules on VeRi-UAV (in \%). Bold and underlined numbers indicate the best and second best ranked results, respectively.}
\end{table}

\begin{figure*}[t]
    \centering
    \includegraphics[width=1\textwidth]{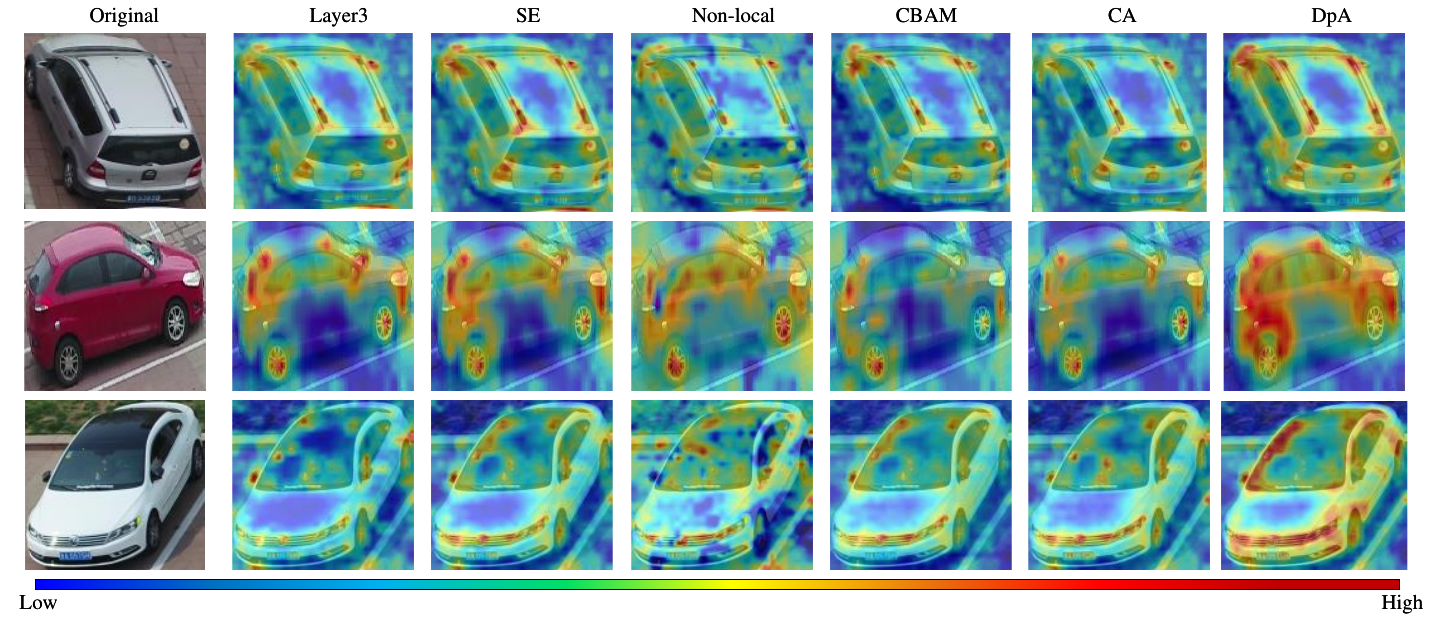}
    \caption{{Heat map comparison of different attention modules. The red area indicates the part of the network with the highest attention value, and the blue area indicates the part of the network with the lowest attention value.
}}
    \label{fig4}
\end{figure*}

To further validate the effectiveness of the DpA method, we also used the Grad-CAM++ technique to visualize the different attention maps. As shown in Figure \ref{fig4}, from left to right, the attention maps of residual layer 3 (without any attention), SE, Non-local, CBAM, CA, and DpA are shown in order. It can be clearly seen that, firstly, all six methods focus on the vehicle itself. secondly, the attention modules of SE, Non-local, CBAM and CA pay less attention to the local information of the vehicle and some important parts are even ignored, while the red area of the DpA module is more obvious to achieve more attention to important cues at different fine-grained levels and to improve the feature extraction capability of the network.

\subsubsection*{Comparison of DpA module placement in the network}
We designed a set of experiments and demonstrated its effectiveness by adding DpA modules at different stages of the backbone network. \checkmark indicates that the DpA module is added after one of the residual blocks of the backbone network.

Table \ref{tab5} shows the experimental results of adding the DpA module after different residual blocks of the backbone. Firstly, the results show that the different residual blocks added to the backbone network have an impact on the network robustness. Specifically, adding the DpA module behind the 2nd (No.1), 3rd (No.2) and 3rd and 4th (No.6) residual blocks of the backbone network respectively improves the accuracy over the baseline (No.0), indicating that the module is able to effectively extract fine-grained vehicle features at these locations. In contrast, adding the DpA module behind the 4th residual block (No.3) and behind the 2nd and 3rd (No.4), 2nd and 4th (No.5) residual blocks all show some decrease in accuracy over the baseline (No.0), which indicates that the network's attention is more dispersed after adding it to these positions, thus introducing more irrelevant information.
Secondly, it can be seen from the table that using mostly one DpA is more robust to the learning of network features than using two DpAs jointly, and saves some training time. In particular, No.2, after adding the DpA module to the third residual block of the backbone network, has at least a 2.17\% improvement in mAP compared to the joint use of two DpA's. In brief, weighing the pros and cons, we choose to add the DpA module only after Conv3\_x of ResNet50. 

\begin{table}[t] 
\centering
\begin{tabular}{|c|c|c|c|c|c|c|c|c|}
\hline
\rowcolor{gray}
\textbf{No.}& \textbf{Conv3\_x}	& \textbf{Conv4\_x} & \textbf{Conv5\_x}& \textbf{mAP} & \textbf{Rank-1} & \textbf{Rank-5} & \textbf{mINP}  & \textbf{Training time} \\
\hline
   0&         &      &     & 79.25       & 95.96        & 98.12   & 49.29   &  \textbf{0.70h}\\
   \hline
   1& \checkmark     &      &    & \underline{79.71}   & 96.23       & 98.21   & \underline{49.35}  & 0.84h\\
   \hline
   2&      & \checkmark     &    & \textbf{81.74}  & \textbf{96.59}   & \textbf{98.48}   &\textbf{51.56}  &  \underline{0.81h}\\
   \hline
   3&      &      & \checkmark     & 78.94   & 96.41      &   98.12  & 48.64  & 1.10h \\
   \hline
   4&    \checkmark  & \checkmark  &  &  78.34      &    96.41   & \underline{98.39} & 46.84  &  1.00h \\
   \hline
   5&     \checkmark  &   & \checkmark  & 78.87     &  96.23   &    97.76   & 47.58    & 1.32h \\
  \hline
   6&     & \checkmark     & \checkmark     & 79.57      &  \underline{96.50}   & \underline{98.39}     & 47.75 &  1.27h\\
   \hline
\end{tabular}
\caption{Ablation experiment of adding DpA module at different residual blocks of the backbone network on VeRi-UAV (in \%). Bold and underlined numbers indicate the best and second best ranked results, respectively.\label{tab5}}
\end{table}

\begin{table}[!t] 
\centering
\begin{tabular}{|l|c|c|c|c|}
 \hline
 \rowcolor{gray}
\textbf{Method}	& \textbf{mAP}	& \textbf{Rank-1} & \textbf{Rank-5}& \textbf{mINP}\\
 \hline
   DpA+Circle~\cite{sun2020circle} &  72.29     & 95.43      &97.67   &38.05    \\ 
 \hline
    DpA+MS~\cite{wang2019multi} & 72.64       &  95.43       &  97.58         &38.31\\
 \hline
    DpA+SupCon~\cite{khosla2020supervised}  &\underline{74.91}      &  \textbf{97.13}    & \underline{98.12}     &\underline{39.04} \\
 \hline
    \textbf{DpA+HMT}        & \textbf{81.74}    & \underline{96.59}     &  \textbf{98.48}    & \textbf{51.56}  \\
 \hline
\end{tabular}
\caption{Ablation experiments of different metric losses on VeRi-UAV (in \%). Bold and underlined numbers indicate the best and second best ranked results, respectively.\label{tab6}}
\end{table}

\subsubsection*{Comparison of different metric losses}
Metric loss has been shown to be effective in Re-ID tasks, which aim to maximize intra-class similarity while minimizing inter-class similarity. The current metric losses treat each instance as an anchor, such as HMT loss and circle loss~\cite{sun2020circle} which utilize the hardest anchor-positive sample pairs. The multi-similarity (MS) loss~\cite{wang2019multi} which selects anchor-positive sample pairs is based on the hardest negative sample pairs. The supervised contrastive (SupCon) loss~\cite{khosla2020supervised} samples all positive samples of each anchor, introducing cluttered triplet while obtaining richer information. The adaptation of different loss functions to the scenario often depends on the characteristics of the training dataset. Table \ref{tab6} shows the experimental results of applying different metric losses for training on the VeRi-UAV dataset, and it can be seen that the HMT loss improves both in terms of mAP compared to other losses, which indicates that the HMT loss targeted to improve the network's ability to discriminate difficult samples for more robust performance in the vehicle Re-ID task in the UAV scenario.

\begin{figure}[t]
\centering
\subfigure[]
{
    \begin{minipage}[b]{.3\linewidth}
        \centering
        \includegraphics[width=5.5cm,height=5.cm]{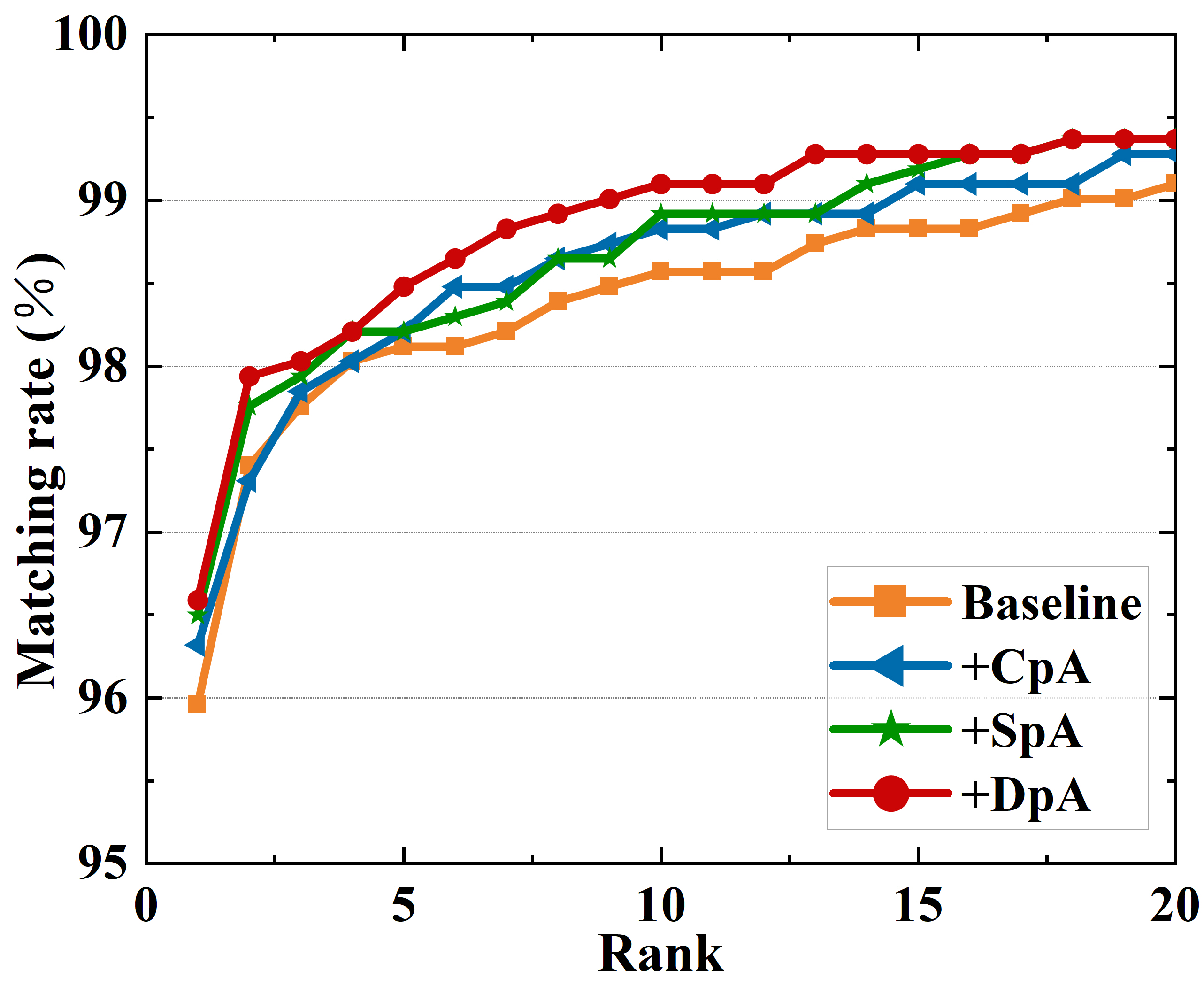}
       
    \end{minipage}
}
\subfigure[]
{
 	\begin{minipage}[b]{.3\linewidth}
        \centering
        \includegraphics[width=5.5cm,height=5.cm]{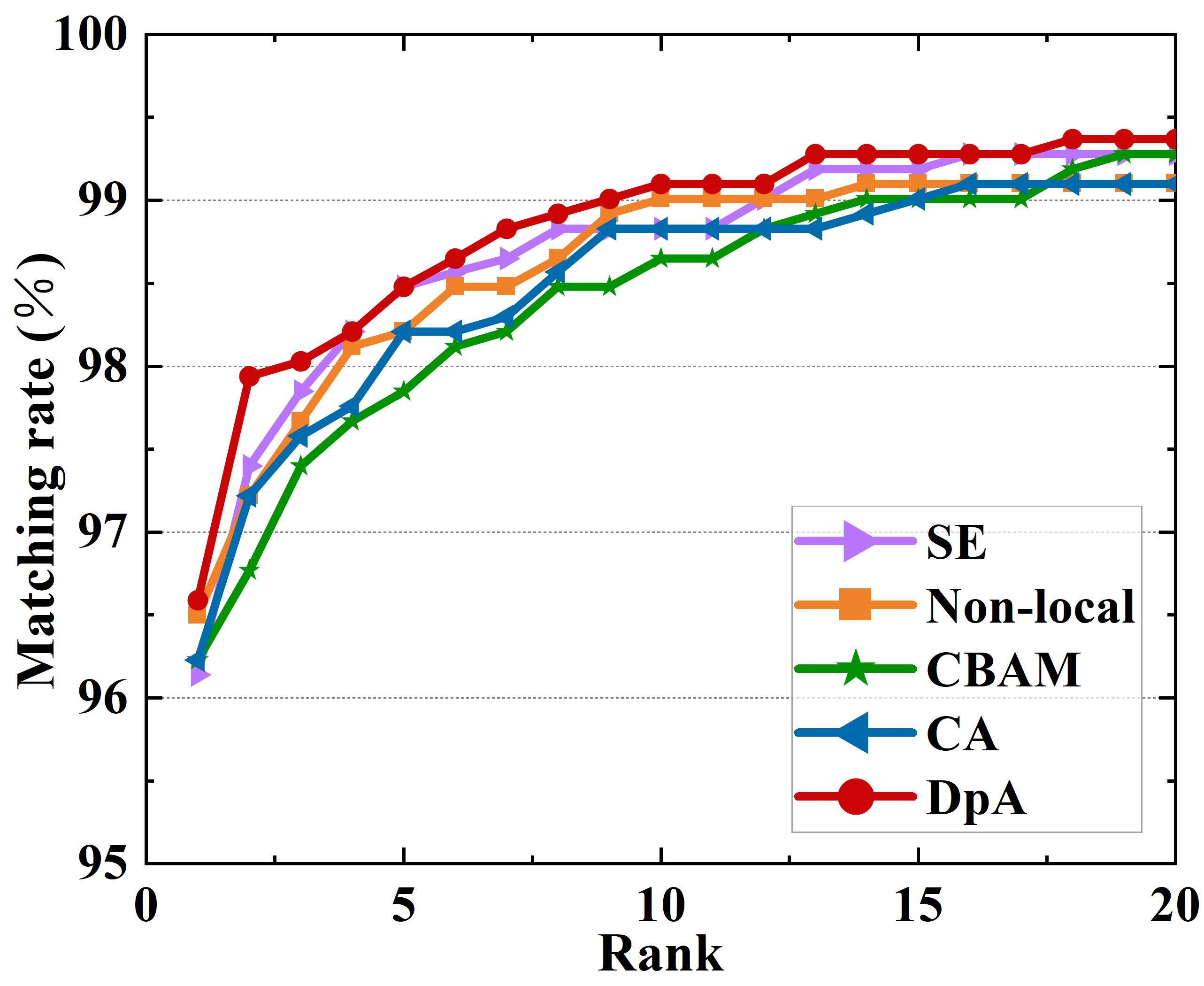}
        
    \end{minipage}
}
\subfigure[]
{
 	\begin{minipage}[b]{.3\linewidth}
        \centering
        \includegraphics[width=5.5cm,height=5.cm]{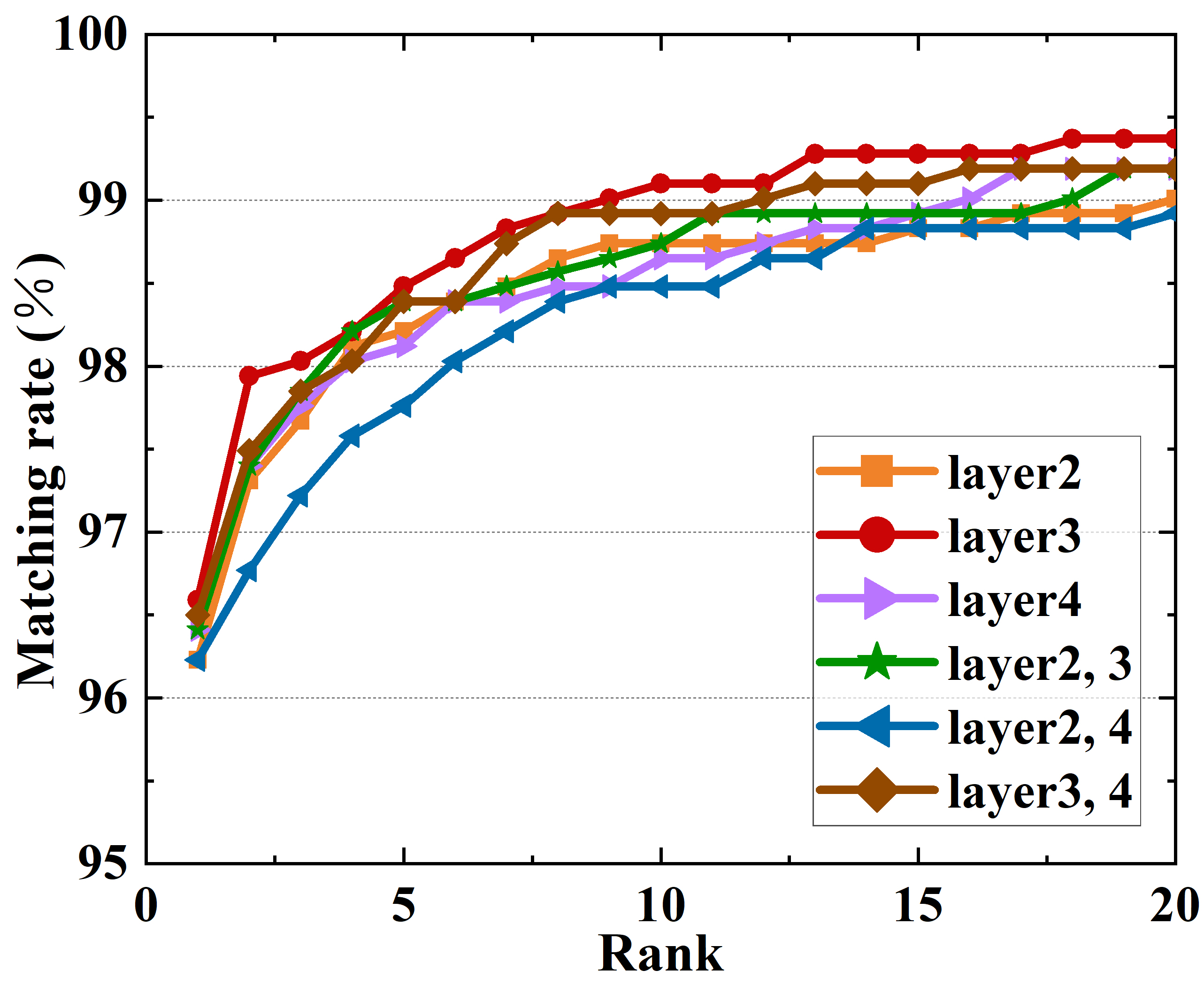}
      
    \end{minipage}
}
\caption{ Comparisons of CMC curves for the case of: (a) CpA, SpA and DpA modules, (b) five different attention mechanisms, and (c) DpA placed in different positions of the backbone network.\label{fig5}}
\end{figure}  

\subsection*{Discussion}
Although the current attention mechanism can achieve certain effect on some vision tasks, its direct application is not effective due to the special characteristics of the UAV shooting angle. Therefore, the main idea of this paper is to design the attention module combining multiple pooling operations and embedding it into the backbone network, which improves the fine-grained information extraction capability for the vehicle Re-ID in the UAV shooting scenario, and devotes to solving the problem of insufficient local information of the vehicle due to the near vertical angle of the UAV shooting and the varying height. In addition, Figure~\ref{fig5} shows the matching rate results from the top 1 to the top 20 for the different validation models mentioned above, respectively. In contrast, the curves plotted by our proposed method as a whole lie above the others, which further validates the effectiveness of the method in terms of actual vehicle retrieval effects, thus providing some support for the injection of UAV technology into intelligent transportation systems.

\subsection*{Visualization of model retrieval results}
\begin{figure}[!t]
\centering
\includegraphics[width=17.8cm]{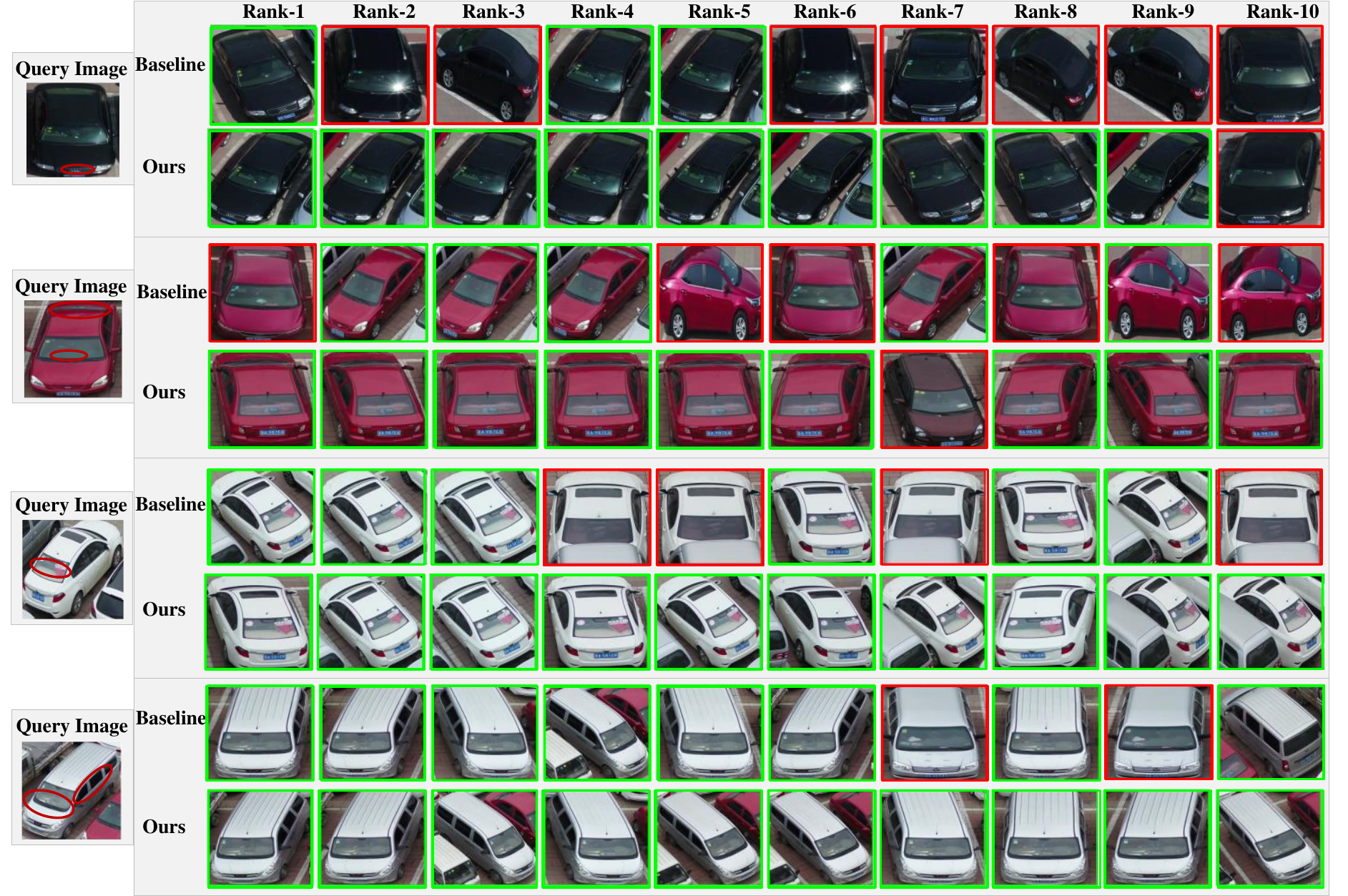}
\caption{Visualization of the ranking lists of model and baseline on VeRi-UAV. For each query, the top and bottom rows show the ranking results for the baseline and joining the DpA module, respectively. The green (red) boxes denote the correct (wrong) results.\label{Vis}}
\end{figure}

To illustrate the superiority of our model more vividly, Figure \ref{Vis} shows the visualization of the top 10 ranked retrieval results for the baseline and model on the VeRi-UAV dataset. A total of four query images corresponding to the retrieval results are randomly shown, the first row for the baseline method and the second row for our method. The images with green borders represent the correct samples retrieved, while the images with red borders are the incorrect samples retrieved.

In contrast, on the one hand, the baseline approach focuses on general appearance features, where the top-ranked negative samples all have similar body postures. However, our method focuses on vehicle features with discriminative information, such as the vehicle parts marked with red circles in the query image in Figure \ref{Vis} (vehicle type symbol, front window, rear window, and side window). On the other hand, as in the second query image in the figure, our method correctly retrieves the top 5 target vehicle samples in only 5 retrievals, while in the baseline method, it takes 9 retrievals to correctly retrieve the top 5 target vehicle samples.

\section*{Conclusion and future work}
In this paper, we propose a dual-pooling attention (DpA) module for vehicle Re-ID to solve the current problem of difficult extraction of local features of vehicles in UAV scenarios due to the high shooting height and vertical shooting angle. The DpA module consists of a channel-pooling attention module and a spatial-pooling attention module, which extracts fine-grained important features of the vehicle by taking two dimensions from the channel domain and the spatial domain. We then fuse the features from both branches to improve the discriminability of the feature representation. Extensive experiments on VeRi-UAV and VRU datasets show that the proposed methodological framework can improve the effective extraction of features from UAV-captured vehicle images and achieve competitive performance in the Re-ID tasks. 

Due to the lack of research on vehicle Re-ID in the UAV aerial photography scene, there is great potential for future research, such as considering expanding UAV scene datasets (e.g., placing drones at different angles to increase the number of vehicle images containing multiple views), combining spatiotemporal information of vehicles, and combining vehicle images captured by fixed surveillance cameras and UAVs for application to vehicle Re-ID tasks.

\bibliography{sample}

\end{document}